\documentclass[letterpaper, 10 pt, conference]{ieeeconf}  

\IEEEoverridecommandlockouts                              

\usepackage{graphics} 
\usepackage{epsfig} 
\usepackage{times} 
\usepackage{amsmath} 
\usepackage{amssymb}  
\usepackage{tikz}
\usepackage{pgfplots}
\usepackage{tikzscale}
\pgfplotsset{compat = 1.3}
\usepackage{siunitx}
\usepackage{hyperref}
\usepackage{siunitx}

\newcommand{\crscarname}{Chronos}
\newcommand{\crsswname}{CRS}

\title{\LARGE \bf
\crscarname~and \crsswname: Design of a miniature car-like robot and a software \\ framework for single and multi-agent robotics and control
}

\author{Andrea Carron$^*$, Sabrina Bodmer, Lukas Vogel,  Ren\'{e} Zurbr\"ugg, David Helm,\\ Rahel Rickenbach,  Simon Muntwiler, Jerome Sieber and  Melanie N. Zeilinger 
\thanks{The authors are members of the Institute of Dynamic Systems and Control, ETH Zurich, Sonneggstrasse~3, 8092 Zurich, Switzerland. The work of Simon Muntwiler was supported by the Bosch Research Foundation im Stifterverband. }
\thanks{$^{*}$Corresponding author {\tt\small carrona@ethz.ch}}%
}

\begin{document}

\maketitle
\thispagestyle{empty}
\pagestyle{empty}

\begin{tikzpicture}[overlay, remember picture]
\node[anchor=south,yshift=8pt] at (current page.south) {\fbox{\parbox{0.99\textwidth}{\footnotesize \textbf{Published in: 2023 IEEE International Conference on Robotics and Automation (ICRA). DOI: 10.1109/ICRA48891.2023.10161434.}\\ 
			\textcopyright \space 2023 IEEE. Personal use of this material is permitted. Permission from IEEE must be obtained for all other uses, in any current or future media, including reprinting/republishing this material for advertising or promotional purposes, creating new collective works, for resale or redistribution to servers or lists, or reuse of any copyrighted component of this work in other works.}}};
\end{tikzpicture}

\begin{abstract}
From both an educational and research point of view, experiments on hardware are a key aspect of robotics and control. In the last decade, many open-source hardware and software frameworks for wheeled robots have been presented, mainly in the form of unicycles and car-like robots, with the goal of making robotics accessible to a wider audience and to support control systems development. Unicycles are usually small and inexpensive, and therefore facilitate experiments in a larger fleet, but they are not suited for high-speed motion. Car-like robots are more agile, but they are usually larger and more expensive, thus requiring more resources in terms of space and money. In order to bridge this gap, we present \crscarname, a new car-like 1/28th scale robot with customized open-source electronics, and \crsswname, an open-source software framework for control and robotics. The \crsswname~software framework includes the implementation of various state-of-the-art algorithms for control, estimation, and multi-agent coordination. With this work, we aim to provide easier access to hardware and reduce the engineering time needed to start new educational and research projects.
\end{abstract}


\section{Introduction}
In robotics and control development, experimental work is key to assess and test the effectiveness of algorithms. Car-like robots are great platforms to achieve this, as they mix agility with dynamic complexity. Furthermore, fleets of car-like vehicles can be used to simulate driving scenarios as well as multi-robot coordination strategies. However, the size and cost of available car-like platforms limit their usage, as they are comparably expensive and large in size. In this paper, we present \crscarname~and \crsswname. The former is an inexpensive, 1/28th scale car-like robot with open-source electronics. The small size of the vehicle enables experiments in small laboratories, as well as running multi-agent experiments in a confined environment. The overall cost of a single agent is below \$$300$, which makes \crscarname~an ideal platform for research and educational projects. 
The \crsswname~framework is an open-source software platform for testing control and robotics applications such as autonomous racing and single and multi-agent robotics. The current implementation comes with algorithms for control, safety, estimation, and multi-agent coordination, which reduces the implementation effort necessary to test new algorithms. The electronics schematics and PCBs, together with the software and firmware stacks are available online and open-source under a BSD 2-clause license in order to foster the wide-spread use of the platform, reduce the setup time, and encourage the exchange of code in the community.

\section{Related Work and Contributions}
A wide range of wheeled robots are available on the market, for this reason, a complete survey of all available platforms is not within the scope of this paper. In this section, we list some of the most popular robots that are/were available in the last three decades and whose design is similar to the platform we propose. We categorize the robots as unicycle- and bicycle-type vehicles~\cite{Corke2013}. The former is a vehicle usually having two parallel-driven wheels, one mounted on each side of the center. The latter includes car-like robots with four wheels, where only the front two are  steerable, usually via an Ackermann steering mechanism.
\begin{figure}[t]
    \centering
    \includegraphics[width=0.8\linewidth]{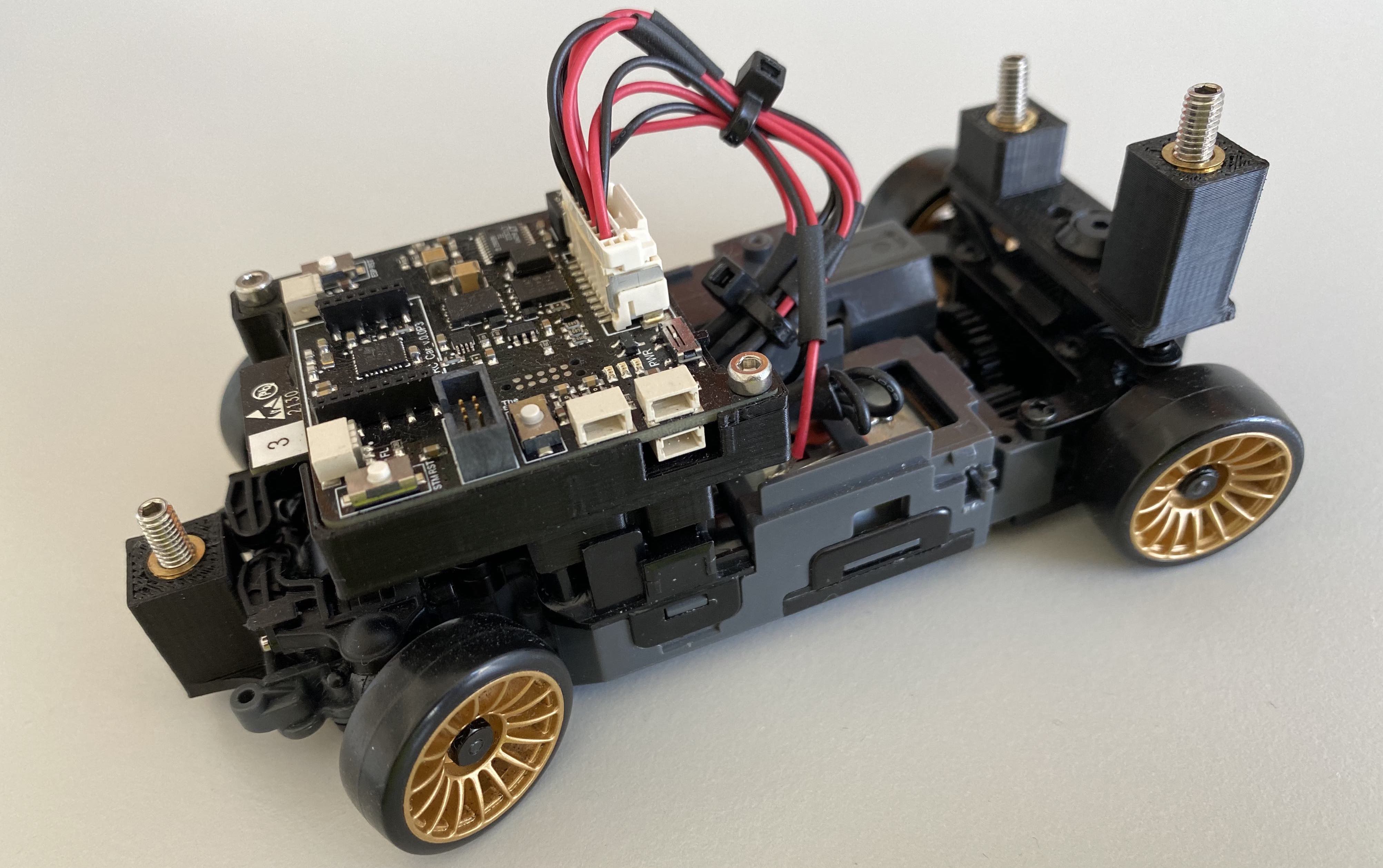}
    \caption{The \crscarname~car. The chassis is taken from a Kyosho Mini-Z miniature car, while the PCB and the 3D printed mounts are custom made. }
    \label{fig:hardware-overview}
\end{figure}
The first successful small-scale unicycle was Khepera~\cite{Mondada1994}. The first generation was designed at EPFL in 1991 and had widespread success in both research~\cite{Verschure2003} and education. Another successful platform was the LEGO Mindstorms RCX~\cite{Lego}, the first robotic platform from LEGO, whose combination with LEGO bricks made it popular for low-budget middle/high-school education~\cite{Kiss2010} and research~\cite{Carron2013} projects. Another popular unicycle, also developed at EPFL, is the e-puck~\cite{mondada:2009}, proposed as a cheaper alternative to the bigger and more expensive Khepera III. 
Over the years, more powerful and expandable unicycle robots became available like iRobot Create~\cite{iRobot}, Pololu 3pi~\cite{Thursk2010}, Swarmrobot~\cite{Kernbach2011},  AERobot~\cite{Rubenstein2015}, GRITSBot~\cite{Pickem2015}, Pheeno~\cite{Wilson2016}, Duckiebots~\cite{Liam2017}, and Wheelbot~\cite{Wheelbot}. More recently, remotely accessible unicycle swarm robotics platforms have been developed, among them, the Mobile Emulab~\cite{Johnson2006}, the HoTDeC testbed~\cite{Stubbs2006}, the Robotarium~\cite{Pickem2017}, and Duckietown~\cite{Liam2017}. A comprehensive list of multi-robot testbeds can be found in~\cite{Jimenez2013}.

While these offer great platforms for  applications related to trajectory planning, their main limitations for high performance controller development are their lack of agility and low maximum speed. This has motivated the use of  car-like vehicles in research and education in recent years. The most common platforms are medium- to full-scale vehicles allowing the integration of powerful computers and to embed various sensors, like IMUs, CMOS/CCD motion sensors, LIDARs, wheel encoders, etc. Among the full-scale prototypes are the Stanford Shelley~\cite{Funke2012}, Stanley~\cite{Thrun2010}, and Marty cars~\cite{Goh2019}; Formula Driverless custom cars~\cite{Kabzan2019} designed to compete in the Formula Student Driverless championship; and most recently Roborace DevBot 2.0~\cite{Roborace} and Dallara AV-21~\cite{Wischnewski2021}. Among the medium-scale vehicles, usually ranging between 1 to 10th and 18th scale, there are the Autorally~\cite{Goldfain2019}, Donkey Car~\cite{Donkeycar}, F1-10 cars~\cite{kelly:2020, Rosolia}, and the AWS Deep Racer~\cite{Deepracer}. The focus of these platforms is mainly oriented towards single-agent autonomous driving/racing and not multi-agent systems, as they require medium- to large-scale infrastructure and are thus comparably expensive and not used for experiments with many agents. The small-scale platforms presented in \cite{Hsieh2006,Leung2007}, and~\cite{Liniger2015} would fit the single- and multi-agent use-case, however, are only partially or not at all open-source, rendering them difficult to recreate.

On the software side, few toolboxes are currently available to solve control tasks. Among them: Drake~\cite{Tedrake2019}, the Control Toolbox~\cite{Giftthaler2018}, which however is not maintained anymore, and OCS2~\cite{OCS2}, which is a toolbox that includes only controllers, but not the full control pipeline including state estimators and safety filters. 

The contribution of this paper is twofold: we present the design of \crscarname, an inexpensive miniature car-like robot with open-source electronics and firmware. Thanks to its small-scale design, the robot can be used in confined spaces. The second contribution is the design of \crsswname, an open-source software framework for performing simulations and experiments. The software aims at supporting both researchers and practitioners by providing a modular framework to design control pipelines for both simulation and hardware experiments. It comes with implementations of state-of-the-art state estimators, controllers, and safety filters, which are non-trivial to set up. Furthermore, we show that \crsswname~allows to quickly go from successful simulations to hardware experiments. Hardware, firmware, and the \crsswname~software framework are available under a BSD clause-2 license\footnote{\url{https://gitlab.ethz.ch/ics/crs}}. Finally, the linked video showcases the platform\footnote{\url{https://youtu.be/l1tMSebViZ0}}. 

\section{Design Principles}
\label{sec:design_philosophy}
In this section, we introduce the over-arching principles according to which \crscarname~and \crsswname~are designed:
\begin{itemize}
    \item \emph{Fast dynamics}: We envision a 4-wheel agile and fast car-like robot that uses an Ackermann steering mechanism. The robot should produce highly dynamic motion, and reach comparably high speeds. 
    \item \emph{Small-scale}: Since space is a limited resource in many laboratories, the robots should be small-scale, allowing for operation in a small area.
    \item \emph{Low-cost}: The cost of a single robot should be low, allowing research groups and students to perform experimental work with limited resources and facilitating the use for multi-agent applications.
    \item \emph{User-friendliness}: The platform should be accessible to people with limited electronics knowledge and intermediate programming skills. This is a key requirement in order to become a widely-used platform for  research and educational projects. 
    \item \emph{Open-source}: Electronics, firmware, and software should be open-source to facilitate their usage.  
\end{itemize}
To the best of the authors' knowledge, there are currently no platforms that satisfy all these criteria, which motivated us to design \crscarname~and \crsswname~in order to bridge this gap.

\section{\crscarname}
\label{sec:car}
In the following, we describe the hardware, electronics, and firmware stack used for~\crscarname.
\begin{figure}[t]
    \centering
    \includegraphics[width=1\linewidth]{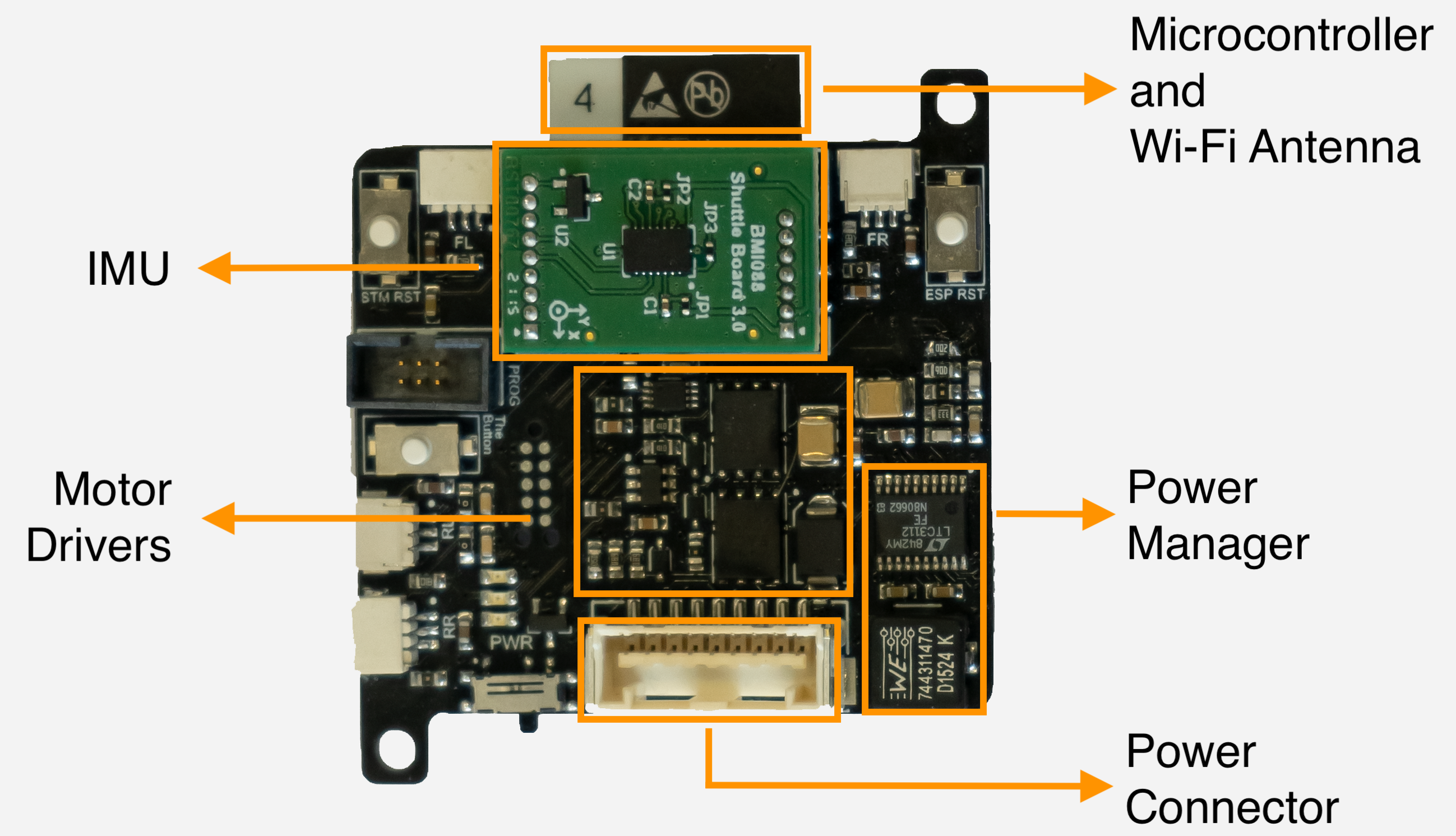}
    \caption{The electronics board of \crscarname. In the picture, the IMU, the motor driver, and the power manager are highlighted. The WiFi antenna is shown, while the microcontroller lies under the IMU.}
    \label{fig:pcb}
\end{figure}

\subsection{Hardware}
The following hardware requirements are derived from the previously stated design principles: reachable speeds of $3-4$ m/s; maximum size of 15 cm x 7 cm, thus allowing for operation in an area of about 3 m x 4 m; and costs below~\$$300$.
Based on these requirements, we choose the Kyosho Mini-z~\cite{Kyosho}, a widely available 1/28th scale RC car, to be the base of \crscarname.  The Kyosho Mini-Z car is equipped with two DC motors, a drive-train and a steering actuator, and a potentiometer acting as a steering encoder. 

We designed a custom electronics board, see Figure~\ref{fig:pcb}, composed of four modules: a microcontroller, a motor driver, sensors, and a power manager. At the core of the electronics board lies an Espressif ESP32-WROOM32D~\cite{esp32}, a 32-bit microcontroller with a built-in antenna module, which supports Wi-Fi connectivity. Additionally, the microcontroller also provides SPI connectivity, which allows the user to connect additional sensors and actuators. 
The motor driver is an STMicroelectronics STSPIN250~\cite{stspin250}, which is used to drive both actuators - the drive-train and the steering actuator. 
The car's sensors include an inertial measurement unit (IMU) and a steering potentiometer. The used IMU is a Bosch BMI088~\cite{boschbmi088}, which is connected to the microcontroller on an SPI bus. 
\crscarname~can be powered with four AAA batteries and it employs a buck-boost converter to maintain a constant supply of $5$V and $2.2$A to the motor. To protect the batteries, we use a low voltage monitoring circuit that turns \crscarname~off when the voltage level descents below a user-defined threshold. 

\subsection{Firmware}
The microcontroller's firmware is based on the ESP-IDF development framework~\cite{espidf}, with FreeRTOS real-time operating system installed on top. The firmware is task-oriented and the main tasks are communication with the~\crsswname~software and the execution of a low-level control loop for the steering angle and the drive-train acceleration. 
\crscarname~and the \crsswname~framework communicate via a local Wi-Fi network. The microcontroller can connect to an access point and receive data through the Internet Protocol (IP). Data is serialized through the usage of Protocol Buffers (Protobuf)~\cite{protobuf} and then sent as User Datagram Protocol (UDP) frames. 

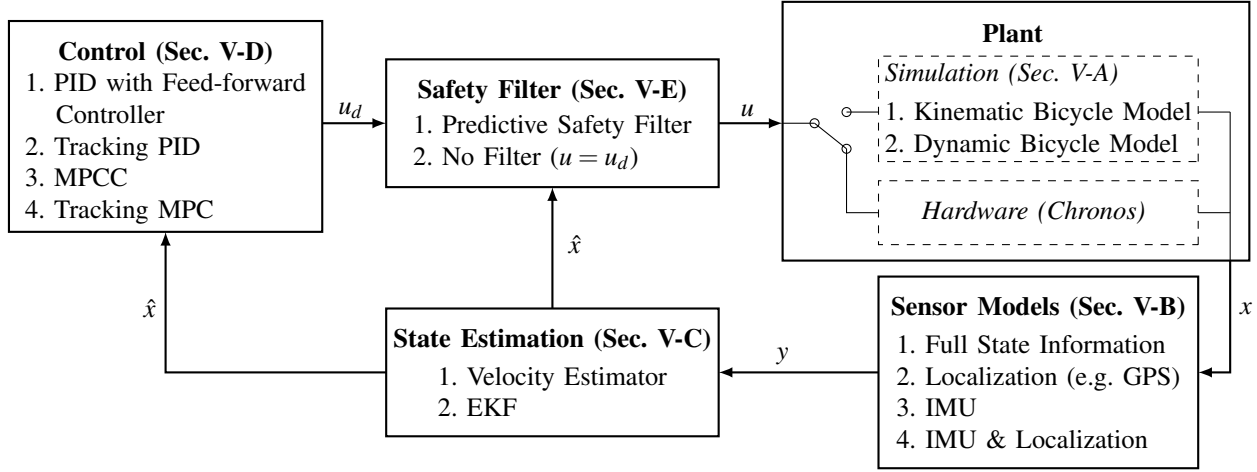
\begin{figure*}[t]
\centering

\begin{tikzpicture}[>=latex,scale = 0.85]

\draw[thick]  (4.8,-3.3) rectangle (-0.1,0);
\node[align = center] at (2.35,-0.5) {\textbf{Control~(Sec.~\ref{subsec:control})}};
\node[align = left] at (2.35,-2.2) {
1.~PID with Feed-forward\\\quad\,Controller\\
2.~Tracking PID\\
3.~MPCC\\
4.~Tracking MPC\\
};

\draw[-latex,thick] (4.8,-1.6) -- (5.8,-1.6);
\node at (5.25,-1.4) {$u_d$};

\draw[thick]  (11,-2.6) rectangle (5.8,-0.6);
\node[align = center] at (8.4,-1.1) {\textbf{Safety Filter~(Sec.~\ref{subsec:SF})}};
\node[align = left] at (8.4,-1.9) {
1.~Predictive Safety Filter\\
2.~No Filter ($u = u_d$)
};

\draw[-latex,thick] (11,-1.6) -- (12,-1.6);
\node at (11.45,-1.4) {$u$};

\draw[thick]  (19.3,-3.75) rectangle (12,0.3);
\node[align = center] at (15.6,-0.2) {\textbf{Plant}};
\draw[dashed]  (18.5,-2.2) rectangle (13.5,-0.55);
\node[align = left] at (16,-1.4) {
\textit{Simulation~(Sec.~\ref{subsec:models})}\\[0.1cm]
1.~Kinematic Bicycle Model\\
2.~Dynamic Bicycle Model
};
\draw[dashed]  (18.5,-3.5) rectangle (13.5,-2.5);
\node[align = left] at (16,-3) {
\textit{Hardware~(\crscarname)}
};
\draw (12,-1.6) -- (12.5,-1.6);
\draw (12.5,-1.6) circle (2pt);
\draw (13,-1.425) -- (13.5,-1.425);
\draw (13,-1.425) circle (2pt);
\draw (13,-2) -- (13,-3) -- (13.5,-3);
\draw (13,-2) circle (2pt);
\draw (12.5,-1.6) -- (13,-2);
\draw (18.5,-1.425) -- (19,-1.425);
\draw (18.5,-3) -- (19,-3);
\draw (19,-1.425) -- (19,-3.75);

\draw[-latex,thick] (19,-3.75) -- (19,-5.5) -- (18.5,-5.5);
\node at (19.25,-4.5) {$x$};

\draw[thick]  (18.5,-7) rectangle (13.5,-4);
\node[align = center] at (16,-4.5) {\textbf{Sensor Models (Sec.~\ref{subsec:sensors})}};
\node[align = left] at (16,-5.8) {
1.~Full State Information\\
2.~Localization (e.g. GPS)\\
3.~IMU\\
4.~IMU \& Localization
};

\draw[-latex,thick] (13.5,-5.5) -- (11,-5.5);
\node at (12,-5.25) {$y$};

\draw[thick]  (11,-6.5) rectangle (5.8,-4.5);
\node[align = center] at (8.4,-5.) {\textbf{State Estimation (Sec.~\ref{subsec:est})}};
\node[align = left] at (8.4,-5.8) {
1.~Velocity Estimator\\
2.~EKF
};

\draw[-latex,thick] (5.8,-5.5) -- (2.35,-5.5) -- (2.35,-3.3);
\node at (2.1,-4.5) {$\hat{x}$};
\draw[-latex,thick] (8.4,-4.5) -- (8.4,-2.6);
\node at (8.75,-3.5) {$\hat{x}$};

\end{tikzpicture}
\caption{Above schematic depicts the control pipeline established by \crsswname,~including options for implemented algorithms described in Section~\ref{sec:current_implementation}. The \crsswname~framework can be used for both simulations and hardware experiments, since the plant block can switch between these two modes.}
\label{fig:architecture} 
\end{figure*}

\section{\crsswname~Software Framework}
\label{sec:current_implementation}
The \crsswname~framework aims to simplify simulation and transition to hardware experiments, as the same framework is used for both simulations and experiments. The framework is divided into two main parts: The first part (\textsl{crs}) contains the implementations of the main modules and it is possible to integrate this part of the framework into other software packages. The modules are in fact designed as toolboxes rather than as integrated applications in order to provide maximum flexibility to the users. Each module is built as a single catkin package enabling the user to compile only the required components. The second part (\textsl{ros4crs}), contains a wrapper around the provided modules, based on the robot operating system (ROS)~\cite{Quigley09} to enable communication among them and allow the user to run \crsswname~in a distributed fashion.

Figure~\ref{fig:architecture} shows a simplified overview\footnote{Two detailed UML diagrams are available at the following links: \url{https://gitlab.ethz.ch/ics/crs/-/wikis/crs.pdf} and \url{https://gitlab.ethz.ch/ics/crs/-/wikis/ros.pdf}} of the modules implemented in~\textsl{crs} and their communication provided by \textsl{ros4crs}. The pipeline consists of a plant, which either represents a simulated dynamics model or real robotics hardware, e.g.,~\crscarname; sensors to (partially) measure the state of the system; a state estimator, which estimates the system state from the sensor readings; a controller, which computes the next control input based on the current state estimate; and a safety filter, which checks if the proposed control input is safe to apply to the system. Given the modular structure of the framework, it is possible to adapt the implementation to custom models/algorithms and build custom control pipelines similar to the one shown in Figure~\ref{fig:architecture}. In the subsequent sections, we summarize the models/algorithms listed in Figure~\ref{fig:architecture} for each module of the control pipeline.

\subsection{Dynamical Models}
\label{subsec:models}
Models are a core component in the \crsswname~framework and are of paramount importance to all elements in the pipeline, from simulation to control and estimation. In the following, we describe two widely used models: The kinematic and the dynamic bicycle models. For ease of presentation, both models are presented in their continuous-time formulation.

\subsubsection{Kinematic Bicycle Model}
The continuous-time kinematic bicycle model~\cite{Rajamani2011} is represented by the following differential equations 
\begin{align}
    \dot{x}_p(t) &= v(t) \cos(\psi(t) + \beta(t)),\\
    \dot{y}_p(t) &= v(t) \sin(\psi(t) + \beta(t)),\\
    \dot{\psi}(t) &= \frac{v(t)}{l_r} \sin(\beta(t)),\\
    \dot{v}(t) &= a(t),
\end{align}
where $x_p$ and $y_p$ are the world frame coordinates of the center of gravity, $\psi$  is the heading angle, $v$ is the total speed of the vehicle, and $\beta$ is the slip angle at the center of mass and is defined as 
$
\beta(t) = \arctan\left(\frac{l_r}{l_r+l_f} \tan\left(\delta(t)\right)\right).
$
The model input $u=[\delta, a]$ consists of the steering angle $\delta$ and the drive-train acceleration $a$. The state is described by the vector $x=[x_p, y_p, \psi, v]$. The model has two parameters, $l_f$ and $l_r$, which represent the distances from the center of mass to the front axle and rear axle, respectively. 

\subsubsection{Dynamic Bicycle Model}
The continuous-time dynamic bicycle model~\cite{Rajamani2011} is governed by the following differential equations
\begin{align}\label{eq:dynamic}
	\begin{aligned}
	\dot{x}_p(t) &= v_x(t) \cos(\psi(t)) - v_y(t) \sin(\psi(t)),\\
	\dot{y}_p(t) &= v_x(t) \sin(\psi(t)) + v_y(t) \cos(\psi(t)),\\
	\dot{\psi}(t) &= \omega(t),\\
	\dot{v}_x(t) &= \frac{1}{m} \big( F_x(t) - F_f(t) \sin(\delta(t)) + mv_y(t) \omega(t)\big), \\
	\dot{v}_y(t) &= \frac{1}{m} \big( F_r(t) + F_f(t) \cos(\delta(t)) - mv_x(t) \omega(t)\big), \\
    \dot{\omega}(t) &= \frac{1}{I_z} \big( F_f(t) l_f \cos(\delta(t)) - F_r(t) l_r\big),
	\end{aligned}
\end{align}
where $m$ is the mass of the vehicle, $I_z$ is the inertia along the $z$-axis, $x_p$, $y_p$, $\psi$ are the x-y coordinates and the yaw angle in the world frame, $v_x$ and $v_y$ are the longitudinal and lateral velocities in the body frame, and $\omega$ is the yaw rate. The lateral tire forces $F_{f}$ and $F_{r}$ are modeled with the simplified Pacejka tire model
\begin{equation}
\label{eq:dynamic_bicycle_lateral_force}
\begin{split}
    \alpha_f(t) &= \arctan\left( \frac{v_y(t) + \omega(t) l_f}{v_x(t)} \right) - \delta(t),\\
    \alpha_r(t) &= \arctan\left( \frac{v_y(t) - \omega(t) l_r}{v_x(t)} \right), \\
    F_{f}(t) &= D_f \sin(C_f \arctan(B_f \alpha_f(t))), \\
    F_{r}(t) &= D_r \sin(C_r \arctan(B_r \alpha_r(t))),
\end{split}
\end{equation}
where $\alpha_f$ and $\alpha_r$ are the front and rear slip angles, and $B_f,B_r,C_f,C_r,D_f,D_r$ are the Pacjeka tire model parameters. 
The longitudinal force is modeled as a single force applied to the center of gravity of the vehicle and is computed as a combination of the drive-train command and the velocity 
\begin{equation}
\label{eq:dynamic_bicycle_longitudinal_force}
F_x(t) = \big(c_1 + c_2v_x(t)\big) a(t) + c_3 v_x^2(t) + c_4,  
\end{equation}
where $c_i$ with $i \in \{1,\ldots,4\}$ are appropriate parameters. 
The input to the system is the same as in the kinematic bicycle model, while the state is $x = [x_p,\ y_p,\ \psi,\ v_x,\ v_y,\ \omega]$.

\subsection{Sensor Models}
\label{subsec:sensors}
Sensor models are important for state estimation and simulation. In the following, we describe two sensor models. 

\subsubsection{Localization Systems}
Localization systems, like, GPS or motion capture systems, usually only provide the pose of the system, i.e., the position and the orientation in the world frame. This results in the discrete-time sensor model
$$
h(x(k)) = 
\begin{bmatrix}
x_p(k)^\top &
y_p(k)^\top &
\psi(k)^\top
\end{bmatrix}^\top\!\!.
$$

\subsubsection{Inertial Measurement Unit}
Inertial measurement units (IMU) provide measurements of linear acceleration and angular velocities. For this reason, this sensor model can only be used with the dynamic bicycle model. The discrete-time measurement equation~$h(x(k))$ is the following
$$
\begin{bmatrix}
\dot{v}_x(k)\\
\dot{v}_y(k)\\
\omega(k)
\end{bmatrix}=\begin{bmatrix}
\frac{1}{m} \big( F_x - F_f \sin(\delta(k)) + mv_y(k) \omega(k)\big) \\
\frac{1}{m}\big(F_r + F_f \cos(\delta(k)) - mv_x(k) \omega(k) \big)\\
\omega(k)
\end{bmatrix},
$$
where the forces can be described using the Pacejka tire model~\eqref{eq:dynamic_bicycle_lateral_force} and the longitudinal force model~\eqref{eq:dynamic_bicycle_longitudinal_force}.

\subsection{State Estimators}
\label{subsec:est}
Most of the available localization systems only provide the spatial location and orientation of the vehicle, while linear and angular velocities have to be estimated. In this section, we present two methods for estimating those quantities: a velocity estimator and an extended Kalman filter. 
\subsubsection{Velocity Estimator}
In the velocity estimator algorithm, the x- and y-velocities in world frame and the yaw-rate are first estimated via finite differences, i.e., 
\begin{align}
v_x^w(k) &= \frac{x_p(k) - x_p(k-1)}{T_s}, \quad
v_y^w(k) = \frac{y_p(k) - y_p(k-1)}{T_s}, \notag \\
\omega(k) &= \frac{\psi(k) - \psi(k-1)}{T_s},
\end{align}
where $T_s$ is the sampling time. 
These estimated quantities can be then filtered using a third-order Butterworth low-pass filter obtaining $\hat{v}_x^{w}(k),\hat{v}_y^{w}(k)$, and $\hat{\omega}(k)$, i.e., the estimated x- and y-velocities in world frame and the yaw-rate. To retrieve the estimate of the linear velocities in body frame the following transformation is performed
\begin{equation}
\begin{split}
&\hat{v}_x(k) = \hat{v}_x^{w}(k) \cos(\hat{\omega}(k)) + \hat{v}_y^{w}(k) \sin(\hat{\omega}(k)),
\\
&\hat{v}_y(k) =   \hat{v}_y^{w}(k) \cos(\hat{\omega}(k)) - \hat{v}_x^{w}(k) \sin(\hat{\omega}(k)).
\end{split}
\end{equation}
\subsubsection{Extended Kalman Filter}
An alternative model-based approach is the extended Kalman filter that exploits a model of the system, e.g., the kinematic or dynamic bicycle model, and a model of the available sensors to estimate the state. The overall system model encompasses two discrete-time equations: the state space model and the output model, where the first model describes the dynamics and the second one the sensor model, i.e., 
\begin{equation}
\begin{split}
    & x(k+1) = f(x(k), u(k)) + w(k),\\
    & y(k) = h(x(k), u(k)) + v(k), 
\end{split}
\end{equation}
where $w(k)$ and $v(k)$ are the process and measurement noise and are usually modeled as Gaussian noise with zero-mean and variance $Q$ and $R$, respectively. The standard extended Kalman filter update and prediction equations~\cite{McGee1985} are implemented.

\subsection{Controllers}
\label{subsec:control}
In the following, we describe three tracking controllers and one controller for autonomous racing, which are natively implemented in the \crsswname~framework. For tracking, the control objective is to accurately track a given reference or path, as e.g., the center line of the track or an optimal racing line. Whereas the objective for autonomous racing is to complete a lap on the track as quickly as possible.

\subsubsection{PID} 
Two PID controllers are currently implemented: a tracking PID that tracks a constant set-point, and a path-following PID that tracks a given path. For the tracking PID, the drive-train and the steering commands depend on the position error $e_p$ and the orientation error $e_{\psi}$, respectively, with the two errors defined as
\begin{align}
    e_{p} &= \sqrt{e_{x}^2 + e_{y}^2}, \qquad
    e_{\psi} = \arctan2(\frac{e_{y}}{e_{x}}) - \psi,
\end{align}
where $e_{x}=x_p-x_r$ and $e_{y} = y_p-y_r$ define the error in x- and y-direction, and $r = (x_{r},y_{r})$ is the set-point. The applied inputs follow from the standard PID controller structure, i.e.,
\begin{align*}
        a(k) &= k_{p,a}e_{p}(k) + k_{i,a}\sum_{j=0}^{k}e_{p}(j) + k_{d,a}(e_{p}(k) - e_{p}(k-1)), \\
        \delta(k) &= k_{p,\delta}e_{\psi}(k) + k_{i,\delta}\!\sum_{j=0}^{k}\!e_{\psi}(j) + k_{d,\delta}(e_{\psi}(k) - e_{\psi}(k\!-\!1)).
\end{align*}

The path-following PID computes the predicted lateral distance to the center line~$e_{l}$ by using the kinematic model, and follows the standard PID controller structure to calculate the steering input.
The applied torque is set to a constant value in order to achieve a desired steady state velocity.

\subsubsection{PID with Feed-forward Controller} The implemented PID controller with feed-forward controller is based on the controller proposed in~\cite{kritayakirana2012} and is intended for path following, e.g., a track's center line or a racing line. It makes use of a lateral feedback controller combined with a feed-forward controller to control the steering angle $\delta$, and a longitudinal feed-forward controller to control the acceleration $a$. 
The lateral controller is designed as in~\cite[Chapter 3]{kritayakirana2012}, where the steering input is computed as
\begin{equation}
    \delta(x) = \delta_{\text{ff}}(x) + \delta_{\text{fb}}(x),
\end{equation}
with feed-forward component $\delta_{\text{ff}}(x)$ computed as
$$
    \delta_{\text{ff}}(x) = \left(l_f + l_r + k_{\text{ug}}(x)\frac{ v_x^\top v_x}{g}\right) \kappa(x),
$$
where $\kappa(x)$ is the curvature of the reference and $k_{\text{ug}}(x)$ is the vehicle under-steer gradient as defined in~\cite[Equation~(3.7)]{kritayakirana2012}.
The lane keeping feedback term $\delta_{\text{fb}}(x)$ is computed as
$$
    \delta_{\text{fb}}(x) = \delta_{\text{control}}(x) + \delta_{\text{damping}}(x),
$$
with
\begin{align*}
    \delta_{\text{control}}(x) = & - k_c \left(e(x) + x_{\text{la}} \sin (\alpha_e(x))\right), \\
    \delta_{\text{damping}}(x) = & - k_d \left(\omega - \|v_x\|_2 \kappa(x) \cos{\alpha_e(x)} \right),
\end{align*}
where $k_c, k_d$ are control gains, $e(x)$ is the lateral error, i.e., the distance of the center of gravity to the closest point on the path, $x_{\text{la}}$ is the look-ahead distance, $\alpha_e(x)$ is the angle error, i.e., the difference between the car orientation and the tangent to the path, and $\omega$ is the yaw rate.

Additionally, a longitudinal controller is used to control the acceleration.
Thereby, the objective is to track a constant target acceleration~$a_{\text{target}}$, which is reduced whenever a turn is ahead.
The heuristic longitudinal control law is computed as
\begin{align}
    a = a_{\text{target}} - k_a\Bar{\kappa}(x)v_x^\top v_x,
\end{align}
where $k_a$ is the control gain and $\Bar{\kappa}(x)$ is the average curvature over a look-ahead horizon.

\subsubsection{Tracking MPC}
The goal of the tracking MPC controller is to track a given reference trajectory or set-point and is formalized as the following nonlinear optimization problem
\begin{align}
    \begin{split}
        \label{eq:tracking_mpc}
        \min_{[u_0, \dots, u_{N}]} &\sum_{i=0}^{N}\|x_i - x^r_i \|_Q^2 + \|u_i-u^r_i\|_R^2  \\
        \text{s.t. } & x_{i+1} = f(x_i, u_i), \ x_0 = x(k), \\
        & x_i \in \mathbb{X}, \; u_i \in \mathbb{U},\\
    \end{split}
\end{align}
where $x^r_i$ and $u^r_i$ are state and input references, the function~$f$ describes the dynamics of the system, using either the discretized kinematic or dynamic bicycle model, and~$\mathbb{X}$ and~$\mathbb{U}$ are state and input constraints. Finally, the integer $N$ is the prediction horizon length.

\begin{figure*}[t]
    \centering
    \includegraphics[width=1\linewidth]{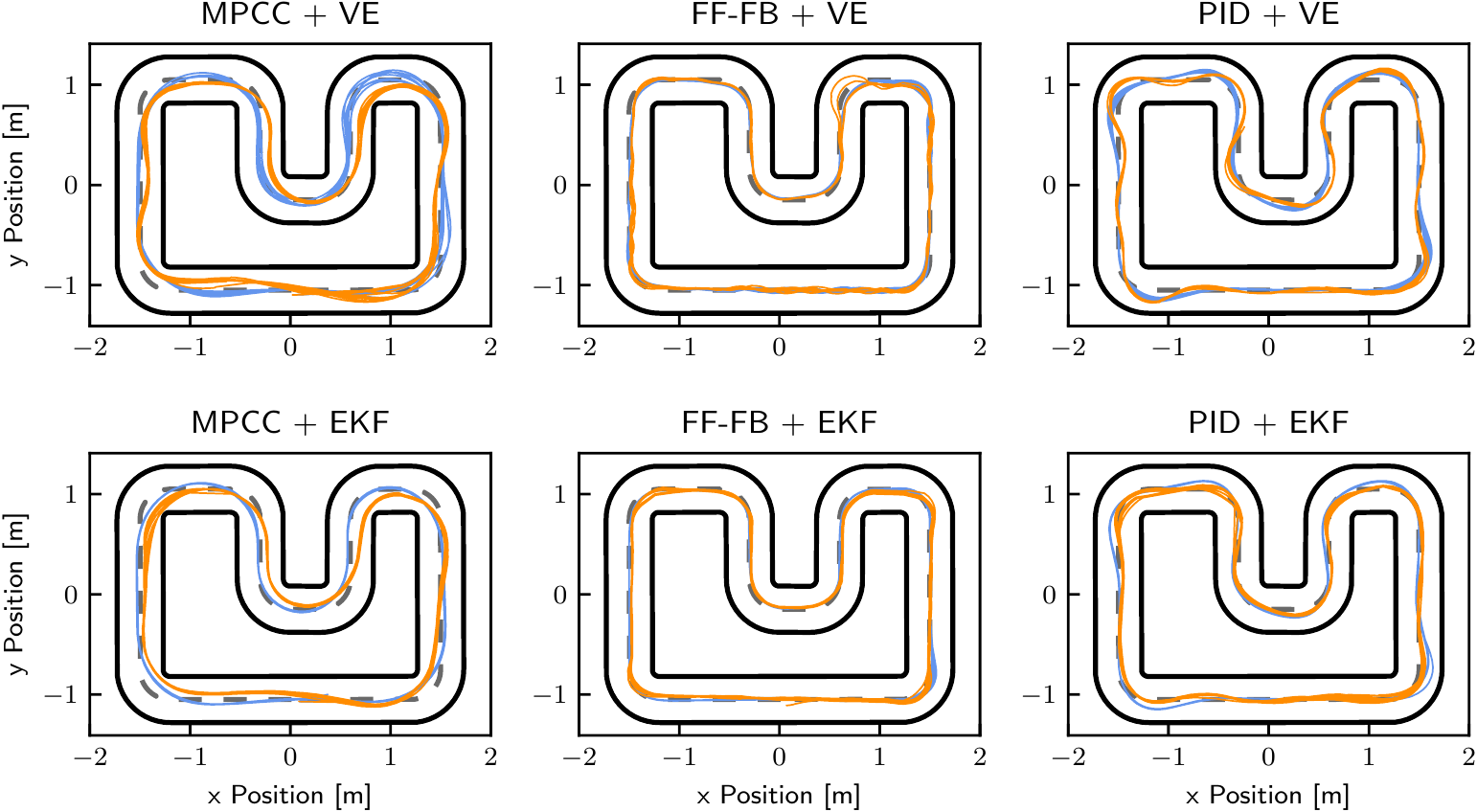}
    \caption{Car trajectories collected from experiments for six control pipelines. The track boundaries are marked in solid black and the center-line in dashed grey. The simulation trajectories are marked in blue and hardware experiments in orange. }
    \label{fig:experiments}
\end{figure*}

\subsubsection{MPCC}
In the following, we briefly review the Model Predictive Contouring Controller (MPCC), one of the most commonly used planning and control algorithms for autonomous racing used for example in~\cite{Liniger2015RCCars,Froehlich2021}. The controller seeks to maximize the progress along a given reference path while satisfying the constraints imposed by the boundaries of the track. The control problem is formalized by the nonlinear optimization problem
\begin{align}
    \begin{split}
        \label{eq:mpcc}
        \min_{[u_0, \dots, u_{N}]} &\sum_{i=0}^{N}\   \| \varepsilon(x_i) \|_Q^2 - Q_{\mathrm{adv}} \gamma(x_i) + \| u_i \|_R^2 \\
        \text{s.t. } & x_{i+1} = f(x_i, u_i), \ x_0 = x(k), \\
        & x_i \in \mathbb{X}, \; u_i \in \mathbb{U}, \gamma_i(x_i) > 0,
    \end{split}
\end{align}
where $\varepsilon(x) = [\varepsilon_l(x), \varepsilon_c(x) ]$ denotes the lag and contouring error, i.e., the lateral and longitudinal error from a given reference path, which depends non-linearly on the state. The variable~$\gamma(x_i)$ represents the progress of the vehicle along a given reference path, e.g., the center line, and is a function of the vehicle's position. The dynamical model~$f$ used in~\eqref{eq:mpcc} can be either the discretized kinematic or dynamic bicycle model described in the previous sections. State and input constraints are defined by the sets $\mathbb{X}$ and $\mathbb{U}$, respectively. Finally, the integer $N$ is the prediction horizon length.

\subsection{Predictive Safety Filter}
\label{subsec:SF}
A predictive safety filter aims to keep a system within state and input constraints despite being controlled by a potentially unsafe control input. The implemented safety filter has a focus on autonomous racing~\cite{Tearle2021}, and guarantees safety by computing a sequence of safe backup control inputs that lead the agent towards a set of known safe states, where the first input of the sequence is as close as possible to the desired control signal $u_d$. The safe control input can be obtained by solving the following nonlinear optimization problem
\begin{align}
    \begin{split}
        \label{eq:tracking_mpc}
        \min_{[u_0, \dots, u_{N}]} & \|u_0 - u_d\|_R^2 + \sum_{i=0}^{N-1} \| \Delta u_{i} \|_{\Delta R}^2\\
        \text{s.t. } & x_{i+1} = f(x_i, u_i), \ x_0 = x(k),  \\
        & x_i \in \mathbb{X}, \ u_i \in \mathbb{U},\ x_N \in \mathbb{X}_N,
    \end{split}
\end{align}
where $\Delta u_{0}:=u_{0}-u(k-1)$, and $\Delta u_{i}:=u_{i}-u_{i-1}$ for $i=1,..,N-1$. The additional term in the cost function is a regularization that penalizes the rate of change of the inputs in order to encourage a smoother control trajectory. State, input, and terminal state constraints are defined by the sets $\mathbb{X}$, $\mathbb{U}$, and $\mathbb{X}_N$, respectively. Finally, the integer $N$ is the prediction horizon length.

\section{Simulations and Experiments}
\label{sec:sim_and_exp}
In this section, we show the behaviour of the following six control pipelines: MPCC, PID with feed-forward controller, and PID, each combined with both the velocity estimator and the EKF as state estimators. 
The trajectories of multiple laps for each of the pipelines tested on simulation (blue) and hardware experiments (orange) are shown in Figure~\ref{fig:experiments}. We initialized both simulations and hardware experiments with same initial conditions. From the figure, we can see that the trajectories obtained in simulation mimic well the ones obtained from hardware experiments. This similarity is key in order to move from successful simulation to hardware experiments rapidly. Figure~\ref{fig:experiments} also shows that using the EKF with any of the presented controllers increases the controller's performance since the time delay introduced in the state estimates is reduced compared to the velocity estimator. Among all presented controllers, the MPCC controller generates the fastest and most aggressive trajectories. The average lap time using the MPCC controller over $10$ laps is $7.23 \si{s} \pm 0.13 \si{s}$ and $6.92 \si{s} \pm 0.11 \si{s}$ obtained with the velocity estimator and EKF, respectively. In this setting, we can generate a highly dynamic motion reaching a maximum speed of about $2.5$~\si{m/s} towards the end of the $2.4$~\si{m} long main-straight and a maximum yaw rate of about $5$~\si{rad/sec} in the two narrower U-turns. Finally, the PID with a feed-forward controller performs better than the PID controller tracking the center line. The predictive capabilities given by the feed-forward term ensure fewer oscillations and higher tracking accuracy.

\section{Conclusions}
\label{sec:conclusions}
In this paper, we presented the design of \crscarname~and~\crsswname. Unlike other car-like platforms, \crscarname~is an inexpensive small-scale robot that can be used for single and multi-agent control and robotics applications in a confined space. 
\crsswname~is a flexible software framework that allows for the easy application of advanced control and estimation algorithms in simulation and hardware experiments. The electronics of \crscarname~and \crsswname~are open-source under a BSD clause-2 license to foster their widespread use.

\section*{Acknowledgement}
The development of~\crscarname~and \crsswname~has been possible thanks to the support of multiple people. We would like to thank Karl Marcus Aaltonen, Christian K\"uttel, Ben Tearle, Daniel Mesham, Robin Fraunenfelder, Marc Rauch, Petar Stamenkovic, Jannes H\"uhnerbein, Alex Hansson, Pascal Sutter, Logan Numerow, and Fenglong Song.

\bibliographystyle{ieeetr}
\bibliography{bibliography}

\begin{thebibliography}{10}

\bibitem{Corke2013}
P.~Corke, {\em Robotics, Vision and Control: Fundamental Algorithms in MATLAB}.
\newblock Springer Publishing Company, Incorporated, 1st~ed., 2013.

\bibitem{Mondada1994}
F.~Mondada, E.~Franzi, and P.~Ienne, ``Mobile robot miniaturisation: A tool for
  investigation in control algorithms,'' in {\em Experimental Robotics III}
  (T.~Yoshikawa and F.~Miyazaki, eds.), (Berlin, Heidelberg), pp.~501--513,
  Springer Berlin Heidelberg, 1994.

\bibitem{Verschure2003}
P.~F. M.~J. Verschure, T.~Voegtlin, and R.~J. Douglas, ``Environmentally
  mediated synergy between perception and behaviour in mobile robots,'' {\em
  Nature}, vol.~425, no.~6958, pp.~620--624, 2003.

\bibitem{Lego}
``Lego mindstorm.'' \url{https://www.lego.com/mindstorms}.
\newblock Accessed: 2022-07-30.

\bibitem{Kiss2010}
G.~Kiss, ``Using the lego-mindstorm kit in german computer science education,''
  in {\em 2010 IEEE 8th International Symposium on Applied Machine Intelligence
  and Informatics (SAMI)}, pp.~101--104, 2010.

\bibitem{Carron2013}
A.~Carron and E.~Franco, ``Receding horizon control of a two-agent system with
  competitive objectives,'' in {\em 2013 American Control Conference},
  pp.~2533--2538, 2013.

\bibitem{mondada:2009}
F.~Mondada, M.~Bonani, X.~Raemy, J.~Pugh, C.~Cianci, A.~Klaptocz, S.~Magnenat,
  J.~christophe Zufferey, D.~Floreano, and A.~Martinoli, ``The e-puck, a robot
  designed for education in engineering,'' in {\em In Proceedings of the 9th
  Conference on Autonomous Robot Systems and Competitions}, pp.~59--65, 2009.

\bibitem{iRobot}
``irobot create.'' \url{https://www.irobot.com/}.
\newblock Accessed: 2022-07-30.

\bibitem{Thursk2010}
B.~Thursk{\'y} and G.~Ga{\v{s}}par, ``Using pololu‘s 3pi robot in the
  education process,'' 2010.

\bibitem{Kernbach2011}
S.~Kernbach, ``Swarmrobot.org - open-hardware microrobotic project for
  large-scale artificial swarms,'' 2011.

\bibitem{Rubenstein2015}
M.~Rubenstein, B.~Cimino, R.~Nagpal, and J.~Werfel, ``Aerobot: An affordable
  one-robot-per-student system for early robotics education,'' in {\em 2015
  IEEE International Conference on Robotics and Automation (ICRA)},
  pp.~6107--6113, 2015.

\bibitem{Pickem2015}
D.~Pickem, M.~Lee, and M.~Egerstedt, ``The gritsbot in its natural habitat - a
  multi-robot testbed,'' in {\em 2015 IEEE International Conference on Robotics
  and Automation (ICRA)}, pp.~4062--4067, 2015.

\bibitem{Wilson2016}
S.~Wilson, R.~Gameros, M.~Sheely, M.~Lin, K.~Dover, R.~Gevorkyan, M.~Haberland,
  A.~Bertozzi, and S.~Berman, ``Pheeno, a versatile swarm robotic research and
  education platform,'' {\em IEEE Robotics and Automation Letters}, vol.~1,
  no.~2, pp.~884--891, 2016.

\bibitem{Liam2017}
L.~Paull, J.~Tani, H.~Ahn, J.~Alonso-Mora, L.~Carlone, M.~Cap, Y.~F. Chen,
  C.~Choi, J.~Dusek, Y.~Fang, D.~Hoehener, S.-Y. Liu, M.~Novitzky, I.~F.
  Okuyama, J.~Pazis, G.~Rosman, V.~Varricchio, H.-C. Wang, D.~Yershov, H.~Zhao,
  M.~Benjamin, C.~Carr, M.~Zuber, S.~Karaman, E.~Frazzoli, D.~Del~Vecchio,
  D.~Rus, J.~How, J.~Leonard, and A.~Censi, ``Duckietown: An open, inexpensive
  and flexible platform for autonomy education and research,'' in {\em 2017
  IEEE International Conference on Robotics and Automation (ICRA)},
  pp.~1497--1504, 2017.

\bibitem{Wheelbot}
A.~R. Geist, J.~Fiene, N.~Tashiro, Z.~Jia, and S.~Trimpe, ``The wheelbot: A
  jumping reaction wheel unicycle,'' {\em IEEE Robotics and Automation
  Letters}, vol.~7, no.~4, pp.~9683--9690, 2022.

\bibitem{Johnson2006}
D.~Johnson, T.~Stack, R.~Fish, D.~M. Flickinger, L.~Stoller, R.~Ricci, and
  J.~Lepreau, ``Mobile emulab: A robotic wireless and sensor network testbed,''
  in {\em Proceedings IEEE INFOCOM 2006. 25TH IEEE International Conference on
  Computer Communications}, pp.~1--12, 2006.

\bibitem{Stubbs2006}
A.~Stubbs, V.~Vladimerou, A.~Fulford, D.~King, J.~Strick, and G.~Dullerud,
  ``Multivehicle systems control over networks: a hovercraft testbed for
  networked and decentralized control,'' {\em IEEE Control Systems Magazine},
  vol.~26, no.~3, pp.~56--69, 2006.

\bibitem{Pickem2017}
D.~Pickem, P.~Glotfelter, L.~Wang, M.~Mote, A.~Ames, E.~Feron, and
  M.~Egerstedt, ``The robotarium: A remotely accessible swarm robotics research
  testbed,'' in {\em 2017 IEEE International Conference on Robotics and
  Automation (ICRA)}, pp.~1699--1706, 2017.

\bibitem{Jimenez2013}
A.~Jim{\'e}nez-Gonz{\'a}lez, J.~R. {Martinez-de Dios}, and A.~Ollero,
  ``Testbeds for ubiquitous robotics: A survey,'' {\em Robotics and Autonomous
  Systems}, vol.~61, no.~12, pp.~1487--1501, 2013.

\bibitem{Funke2012}
J.~Funke, P.~Theodosis, R.~Hindiyeh, G.~Stanek, K.~Kritatakirana, C.~Gerdes,
  D.~Langer, M.~Hernandez, B.~Müller-Bessler, and B.~Huhnke, ``Up to the
  limits: Autonomous audi tts,'' in {\em 2012 IEEE Intelligent Vehicles
  Symposium}, pp.~541--547, 2012.

\bibitem{Thrun2010}
S.~Thrun, ``Toward robotic cars,'' {\em Commun. ACM}, vol.~53, p.~99–106, apr
  2010.

\bibitem{Goh2019}
J.~Y. Goh, T.~Goel, and J.~Christian~Gerdes, ``{Toward Automated Vehicle
  Control Beyond the Stability Limits: Drifting Along a General Path},'' {\em
  Journal of Dynamic Systems, Measurement, and Control}, vol.~142, 11 2019.
\newblock 021004.

\bibitem{Kabzan2019}
J.~Kabzan, M.~Valls, V.~Reijgwart, H.~Hendrikx, C.~Ehmke, M.~Prajapat,
  A.~B{\"u}hler, N.~Gosala, M.~Gupta, R.~Sivanesan, A.~Dhall, E.~Chisari,
  N.~Karnchanachari, S.~Brits, M.~Dangel, I.~Sa, R.~Dube, A.~Gawel,
  M.~Pfeiffer, and R.~Siegwart, ``Amz driverless: The full autonomous racing
  system,'' 05 2019.

\bibitem{Roborace}
``Roborace.'' \url{https://roborace.com}.
\newblock Accessed: 2022-07-30.

\bibitem{Wischnewski2021}
A.~Wischnewski, M.~Geisslinger, J.~Betz, T.~Betz, F.~Fent, A.~Heilmeier,
  L.~Hermansdorfer, T.~Herrmann, S.~Huch, P.~Karle, F.~Nobis, L.~{\"O}gretmen,
  M.~Rowold, F.~Sauerbeck, T.~Stahl, R.~Trauth, M.~Lienkamp, and B.~Lohmann,
  ``Indy autonomous challenge - autonomous race cars at the handling limits,''
  in {\em 12th International Munich Chassis Symposium 2021} (P.~Pfeffer, ed.),
  (Berlin, Heidelberg), pp.~163--182, Springer Berlin Heidelberg, 2022.

\bibitem{Goldfain2019}
B.~Goldfain, P.~Drews, C.~You, M.~Barulic, O.~D. Velev, P.~Tsiotras, and J.~M.
  Rehg, ``Autorally: An open platform for aggressive autonomous driving,'' {\em
  IEEE Control Systems}, vol.~39, pp.~26--55, 2019.

\bibitem{Donkeycar}
``Donkey car.'' \url{http://www.donkeycar.com}.
\newblock Accessed: 2022-07-30.

\bibitem{kelly:2020}
M.~O'Kelly, H.~Zheng, D.~Karthik, and R.~Mangharam, ``F1tenth: An open-source
  evaluation environment for continuous control and reinforcement learning,''
  in {\em Proceedings of the NeurIPS 2019 Competition and Demonstration Track}
  (H.~J. Escalante and R.~Hadsell, eds.), vol.~123 of {\em Proceedings of
  Machine Learning Research}, pp.~77--89, PMLR, 08--14 Dec 2020.

\bibitem{Rosolia}
G.~M. Jon~Gonzales, Ugo~Rosolia, ``Barc: Berkeley autonomous race car
  project.''

\bibitem{Deepracer}
``Aws deep racer.'' \url{https://docs.aws.amazon.com/deepracer/}.
\newblock Accessed: 2022-07-30.

\bibitem{Hsieh2006}
C.~Hsieh, Y.-L. Chuang, Y.~Huang, K.~Leung, A.~Bertozzi, and E.~Frazzoli, ``An
  economical micro-car testbed for validation of cooperative control
  strategies,'' in {\em 2006 American Control Conference}, pp.~6 pp.--, 2006.

\bibitem{Leung2007}
K.~K. Leung, C.~H. Hsieh, Y.~R. Huang, A.~Joshi, V.~Voroninski, and A.~L.
  Bertozzi, ``A second generation micro-vehicle testbed for cooperative control
  and sensing strategies,'' in {\em 2007 American Control Conference},
  pp.~1900--1907, 2007.

\bibitem{Liniger2015}
A.~Liniger, A.~Domahidi, and M.~Morari, ``Optimization-based autonomous racing
  of 1:43 scale rc cars,'' {\em Optimal Control Applications and Methods},
  vol.~36, no.~5, pp.~628--647, 2015.

\bibitem{Tedrake2019}
R.~Tedrake and the Drake Development~Team, ``Drake: Model-based design and
  verification for robotics,'' 2019.

\bibitem{Giftthaler2018}
M.~Giftthaler, M.~Neunert, M.~St{\"a}uble, and J.~Buchli, ``The control toolbox
  --- an open-source c++ library for robotics, optimal and model predictive
  control,'' in {\em 2018 IEEE International Conference on Simulation,
  Modeling, and Programming for Autonomous Robots (SIMPAR)} (H.~Kurniawati,
  E.~Drumwright, B.~MacDonald, T.~Fraichard, and N.~Ye, eds.), (Piscataway,
  NJ), pp.~123 -- 129, IEEE, 2018-06-11.

\bibitem{OCS2}
R.~S. L.~E. Zurich, ``Ocs2: Optimal control for switched systems,'' 2022.

\bibitem{Kyosho}
``Kyosho.'' \url{http://kyosho.com/mini-z-info/}.
\newblock Accessed: 2022-08-12.

\bibitem{esp32}
``Espressif {ESP32 WROOM} {datasheet}, howpublished =
  {\url{https://www.espressif.com/sites/default/files/documentation/esp32-wroom-32d_esp32-wroom-32u_datasheet_en.pdf}},
  note = {Accessed: 2022-08-12}.''

\bibitem{stspin250}
``{STMicroelectronics} {STSPIN250} datasheet.''
  \url{https://www.st.com/resource/en/datasheet/stspin250.pdf}.
\newblock Accessed: 2022-08-12.

\bibitem{boschbmi088}
``{B}osch {BMI088} datasheet.''
  \url{https://download.mikroe.com/documents/datasheets/BMI088_Datasheet.pdf}.
\newblock Accessed: 2022-08-12.

\bibitem{espidf}
``{ESP-IDF} development framework.''
  \url{https://docs.espressif.com/projects/esp-idf/en/latest/esp32/get-started/index.html}.
\newblock Accessed: 2022-08-12.

\bibitem{protobuf}
Google, ``Protocol buffers.''
  \url{http://code.google.com/apis/protocolbuffers/}.

\bibitem{Quigley09}
M.~Quigley, B.~Gerkey, K.~Conley, J.~Faust, T.~Foote, J.~Leibs, E.~Berger,
  R.~Wheeler, and A.~Ng, ``{ROS}: an open-source robot operating system,'' in
  {\em Proc. of the IEEE Intl. Conf. on Robotics and Automation (ICRA) Workshop
  on Open Source Robotics}, 2009.

\bibitem{Rajamani2011}
R.~Rajamani, {\em Vehicle Dynamics and Control}.
\newblock Mechanical Engineering Series, Springer US, 2011.

\bibitem{McGee1985}
L.~A. McGee, S.~F. Schmidt, and A.~R. Center., {\em Discovery of the Kalman
  filter as a practical tool for aerospace and industry [microform] / Leonard
  A. McGee and Stanley F. Schmidt}.
\newblock National Aeronautics and Space Administration, Ames Research Center
  Moffett Field, Calif, 1985.

\bibitem{kritayakirana2012}
K.~M. Kritayakirana, {\em Autonomous vehicle control at the limits of
  handling}.
\newblock {P}h{D} thesis, 2012.

\bibitem{Liniger2015RCCars}
A.~Liniger, A.~Domahidi, and M.~Morari, ``Optimization-based autonomous racing
  of 1: 43 scale rc cars,'' vol.~36, no.~5, pp.~628--647, 2015.

\bibitem{Froehlich2021}
L.~P. Fröhlich, C.~Küttel, E.~Arcari, L.~Hewing, M.~N. Zeilinger, and
  A.~Carron, ``Model learning and contextual controller tuning for autonomous
  racing,'' 2021.

\bibitem{Tearle2021}
B.~Tearle, K.~P. Wabersich, A.~Carron, and M.~N. Zeilinger, ``A predictive
  safety filter for learning-based racing control,'' {\em IEEE Robotics and
  Automation Letters}, vol.~6, no.~4, pp.~7635--7642, 2021.

\end{thebibliography}

\end{document}